\documentclass[conference]{IEEEtran}
\IEEEoverridecommandlockouts  

\usepackage{cite}
\usepackage{amsmath,amssymb,amsfonts}
\usepackage{algorithm}
\usepackage{algorithmic}
\usepackage{graphicx}
\usepackage{textcomp}
\usepackage{xcolor}
\usepackage{listings}
\usepackage{caption}
\usepackage{float}

\begin{document}

\title{Advanced Image Generation: Negative Prompt Optimization and Latent Classifier Guidance\\
\thanks{}}

\author{%
    \IEEEauthorblockN{Vaddi Charan Sai Nandan Reddy, Harini B, Chandana M S}
    \IEEEauthorblockA{Department of Computer Science, PES University, Bangalore, India \\
    Email: {pes1202202250@pesu.pes.edu}}
}

\maketitle

\begin{abstract}
We present a novel system that integrates negative prompt optimization via a fine-tuned sequence-to-sequence LLM and latent-space classifier guidance to improve the quality of images generated by Stable Diffusion. Our approach automatically generates optimized negative prompts, and employs a CNN–RNN hybrid classifier to evaluate and guide diffusion steps, rolling back low-quality latent updates. Experimental results demonstrate that our dual-guidance framework reduces artifacts and improves semantic fidelity compared to baseline diffusion. Code and demonstration are released under an open-source license.
\end{abstract}

\begin{IEEEkeywords}
Stable Diffusion, Negative Prompt Optimization, Latent Classifier Guidance, Streamlit, CNN–RNN, Generative Models
\end{IEEEkeywords}

\section{Introduction}
Diffusion models such as Stable Diffusion have revolutionized text-to-image generation. However, they still suffer from artifacts, semantic drift, and failure to adhere to user intent. Recent work in negative prompt engineering and classifier guidance addresses these challenges separately. In this paper, we combine both techniques into a unified framework implemented in a user-friendly Streamlit application. Our contributions are:
\begin{itemize}
  \item Automatic negative prompt generation using a fine-tuned Seq2Seq LLM.
  \item A novel ImprovedLatentCNN\_RNN\_Classifier that evaluates intermediate latents and guides diffusion by rolling back undesirable steps.
  \item Empirical evaluation demonstrating improved image quality and reduced artifacts.
\end{itemize}

\section{Literature Review}
\subsection{Optimizing Negative Prompts for Enhanced Aesthetics and Fidelity in Text-To-Image Generation}

The paper titled \textit{Optimizing Negative Prompts for Enhanced Aesthetics and Fidelity in Text-To-Image Generation} proposes \textbf{NegOpt}, a novel two-phase approach for generating effective negative prompts to improve image quality in diffusion models. Traditionally hand-crafted negative prompts are labor-intensive, but NegOpt automates this using a  dataset \texttt{mikeogezi/negopt\_full}, consisting of 256,224 samples. Two key subsets are used: a Supervised Fine Tuning (SFT) dataset with 5,790 high-like samples, and a Reinforcement Learning (RL) subset of 466 samples.

In the SFT phase, a seq2seq model is fine-tuned on prompt–negative prompt pairs using the Adam optimizer with a learning rate of $1e^{-3}$ over 16 epochs. The RL phase further optimizes using scalar rewards based on image quality metrics, with Adam optimizer at $1e^{-5}$ for 8 epochs.

Evaluation metrics include Inception Score (IS), CLIP Score, Aesthetics Score, and Human Evaluation, showing that NegOpt enhances both semantic alignment and visual appeal of generated images. This study highlights the potential of automated negative prompt generation in improving generative model outputs.

\subsection{INITNO: Boosting Text-to-Image Diffusion Models via Initial Noise Optimizatio}

The paper \textit{INITNO: Boosting text-to-image diffusion models by initial noise optimization} introduces a method to improve text-image alignment by optimizing the initial noise input to diffusion models. Unlike traditional methods that optimize during the denoising process, INITNO focuses on pre-selecting high-quality noise samples using attention-based metrics.

The methodology uses two novel metrics: \textbf{Cross-Attention Response Score}, which measures attention to text tokens, and \textbf{Self-Attention Conflict Score}, which detects subject overlap or mixing. Based on these, the latent noise space is partitioned into valid and invalid regions.

The optimization process involves adjusting the noise sample using a joint loss that combines cross-attention response loss, self-attention conflict loss, and distribution alignment loss. This ensures accurate attribute binding without deviating from the standard Gaussian noise distribution.

INITNO is evaluated on structured datasets (Animal-Animal, Animal-Object, Object-Object), and demonstrates superior qualitative and quantitative performance over baselines such as Attend-and-Excite, A-STAR. It improves CLIP similarity scores and achieves 63.33\% preference in a user study, though at the cost of increased inference time (18.93s per image).

\subsection{Fine-Grained Alignment and Noise Refinement for Compositional Text-to-Image Generation}

The paper \textit{Fine-Grained Alignment and Noise Refinement for Compositional Text-to-Image Generation} addresses challenges in capturing intricate prompt details, including missing entities, incorrect attribute bindings, and spatial misalignments in text-to-image diffusion models. It introduces a dual focus on alignment and noise refinement to improve compositional fidelity.

The model is evaluated on two benchmarks: \textbf{T2I-CompBench}, targeting compositional text-to-image evaluation, and the \textbf{HRS Benchmark}, which assesses generation involving three-entity relationships.

The methodology comprises three branches: spatial layout guidance using bounding boxes, prompt decomposition via Large Language Models (LLMs), and attention map optimization. A key contribution is the \textbf{Entity-Attribute-Relation (EAR) Loss}, combining entity mixing and missing loss, attribute binding loss, and spatial relation loss to enforce fine-grained alignment.

Additionally, the paper proposes a \textbf{Fine-Grained Initial Noise Refinement} strategy. A verifier system detects alignment failures post-generation, prompting \textbf{targeted noise correction} rather than full regeneration. This feedback-driven approach enhances alignment in subsequent image generation cycles.

Experimental results confirm improvements over prior models in multi-entity and compositional image generation tasks, demonstrating the effectiveness of both EAR loss and localized noise refinement.

\subsection{Controllable Generation with Text-to-Image Diffusion Models: A Survey}

The paper \textit{Controllable Generation with Text-to-Image Diffusion Models: A Survey} explores recent advancements in text-to-image diffusion models, emphasizing the need for controllability in generation outputs. It reviews mechanisms such as negative prompts and noise refinement, that improve control over the generation process.

The study utilizes two datasets for evaluation: \textbf{ImageNet-100}, consisting of 130,000 images across 100 classes, and \textbf{CIFAR-10}, with 60,000 images across 10 classes. Standard preprocessing techniques, including resizing, normalization, and horizontal flipping, are applied.

The paper categorizes diffusion approaches into three major types: \textbf{Denoising Diffusion Probabilistic Models (DDPM)}, \textbf{Denoising Diffusion Implicit Models (DDIM)} for accelerated generation, and \textbf{Latent Diffusion Models (LDM)} for efficient high-resolution synthesis. It also discusses \textbf{Classifier-Free Guidance (CFG)} as a method to improve prompt-image alignment without requiring class labels.

Experiments conducted on CIFAR-10 and ImageNet-100 assess the performance and efficiency of different noise scheduling techniques, such as linear and cosine schedules. Evaluation metrics include the \textbf{Inception Score (IS)} to evaluate image quality and diversity, and the \textbf{Fréchet Inception Distance (FID)} to measure realism.

Key findings highlight that DDIM provides a 4× speedup over DDPM while preserving quality, cosine noise scheduling enhances sample stability, and LDMs deliver high resolution results at reduced computational cost.

\section{Proposed Methodology}
In this section, we provide an in-depth description of each component, design rationales, and algorithmic procedures. We break down the methodology into four stages: (1) architecture overview, (2) negative prompt optimization, (3) diffusion with DDIM scheduling, and (4) latent classifier guidance. Pseudocode for the full generation loop is given in Algorithm~\ref{alg:full}.

\subsection{Architecture Overview}
Our system integrates three modules in a pipeline):
\begin{enumerate}
  \item \textbf{Negative Prompt Generator}: Produces a prompt $p^{-}$ that suppresses unwanted features.
  \item \textbf{Diffusion Engine}: A Stable Diffusion pipeline with DDIM scheduler generates and updates latent $z_{t}$.
  \item \textbf{Latent Classifier}: A CNN–RNN hybrid evaluates latent sequences and decides whether to accept or rollback each update.
\end{enumerate}
This modular design allows independent training and easy ablation studies.

\subsection{Negative Prompt Optimization}
We model negative prompt generation as a conditional language generation task. Let $p^{+}=\{w_{1},...,w_{n}\}$ be the input tokens. We fine-tune a pre-trained Seq2Seq model $f_{\theta}$ (T5-base) on pairs $(p^{+}, p^{-}_{\mathrm{gt}})$ with the cross-entropy objective:
\begin{equation}
  \mathcal{L}_{\mathrm{neg}} = -\sum_{t=1}^{m} \log P_{\theta}(y_{t} \mid y_{<t}, p^{+}),
\end{equation}
where $y_{t}$ are tokens of the ground-truth negative prompt of length $m$. During inference, we perform beam search:
\begin{algorithm}[H]
  \caption{Negative Prompt Beam Search}
  \label{alg:beam}
  \begin{algorithmic}[1]
    \STATE \textbf{Input:} positive prompt $p^{+}$, beam width $B$, max length $L_{\max}$
    \STATE Initialize beams with start token
    \FOR{$t=1$ to $L_{\max}$}
      \STATE Expand each beam by all vocabulary tokens
      \STATE Score new beams by log-probability
      \STATE Keep top-$B$ beams
      \STATE If all beams end with end-of-sequence token, break
    \ENDFOR
    \STATE \textbf{Output:} highest-scoring beam as $p^{-}$
  \end{algorithmic}
\end{algorithm}
This yields diverse and effective negative prompts in interactive time.

\subsection{Diffusion with DDIM Scheduler}
We adopt the DDIM update rule for deterministic sampling \cite{song2020denoising}. Let $x_{T}\sim\mathcal{N}(0,I)$ and text embeddings $e$ from the prompt pair $(p^{+},p^{-})$. The UNet predicts noise $\epsilon_{\theta}(x_{t}, t, e)$. Then:
\begin{equation}
  x_{t-1} = \sqrt{\alpha_{t-1}}\left(\frac{x_{t} - \sqrt{1-\alpha_{t}}\,\epsilon_{\theta}}{\sqrt{\alpha_{t}}}\right) + \sqrt{1-\alpha_{t-1}}\,\epsilon_{\theta}.
\end{equation}
Here $\{\alpha_{t}\}$ are precomputed schedule parameters. We omit noise injection for deterministic paths.

\subsection{Latent Classifier Guidance}
We propose a two-stage classifier $g_{\phi}=h\circ c$ where:
\begin{itemize}
  \item $c$: CNN encoder extracts spatial features from each 4-channel latent $z_{t}$, producing feature vector $f_{t}\in\mathbb{R}^{128}$.
  \item $h$: a bidirectional GRU aggregates $\{f_{t-k+1},...,f_{t}\}$ into hidden state $h_{t}\in\mathbb{R}^{512}$, followed by an MLP to produce logit $\ell_{t}$.
\end{itemize}
The acceptance probability is $s_{t}=\sigma(\ell_{t})$. We set threshold $\tau=0.5$ and allow up to $R=5$ rollbacks. Algorithm~\ref{alg:full} integrates this into the diffusion loop.

\begin{algorithm}[H]
  \caption{Dual-Guidance Diffusion Generation}
  \label{alg:full}
  \begin{algorithmic}[1]
    \STATE \textbf{Input:} $p^{+}, p^{-}, T, \tau, R$
    \STATE Initialize $z_{T}\sim\mathcal{N}(0,I)$, rollback\_count$=0$
    \FOR{$t=T$ down to $1$}
      \STATE $\hat{z}_{t-1}$ = DDIM\_update$(z_{t},t,p^{+},p^{-})$
      \STATE Append $\hat{z}_{t-1}$ to buffer
      \IF{buffer length $\ge k$}
        \STATE Compute $s_{t-1}=g_{\phi}$(recent $k$ latents)
        \IF{$s_{t-1}<\tau$ and rollback\_count$<R$}
          \STATE $z_{t-1}=z_{t-1}^{\mathrm{acc}}$ (last accepted)
          \STATE rollback\_count++
          \STATE \textbf{continue}
        \ENDIF
      \ENDIF
      \STATE Accept: $z_{t-1}^{\mathrm{acc}}=\hat{z}_{t-1}$, rollback\_count=0
      \STATE $z_{t-1}=\hat{z}_{t-1}$
    \ENDFOR
    \STATE \textbf{Output:} $x_{0}=\text{VAE.decode}(z_{0})$
  \end{algorithmic}
\end{algorithm}

This discrete decision mechanism helps avoid cumulative errors by reverting undesirable updates.

\section{Latent Classifier Performance}
We evaluated our ImprovedLatentCNN\_RNN\_Classifier and several benchmark models on a held-out set of 200 labeled latent sequences (10-step sequences), representing 20\% of our synthetic latent-sequence dataset. Table~\ref{tab:performance} reports standard classification metrics for the ImprovedLatentCNN\_RNN\_Classifier on this set, showing precision, recall, F1‐score, and support per class. Overall performance metrics, including accuracy, ROC AUC, and Brier score, are also reported.

\begin{table}[htbp]
    \caption{ImprovedLatentCNN RNN Classifier performance on binary accept/reject labels.}
    \label{tab:performance}
    \centering
    \begin{tabular}{|l|c|c|c|c|}
        \hline
        Class & Precision & Recall & F1-score & Support \\
        \hline
        0 (reject) & 0.4200 & 0.4565 & 0.4375 & 46 \\
        1 (accept) & 0.5000 & 0.4630 & 0.4808 & 54 \\
        \hline
    \end{tabular}
\end{table}

Overall classifier discrimination and calibration for the ImprovedLatentCNN\_RNN\_Classifier are quantified by:
\[
  \text{ROC AUC} = 0.4642, 
  \quad \text{Brier score} = 0.3121.
\]
These results, particularly the AUC below 0.5, indicate that the classifier performs worse than random guessing on this task, suggesting significant room for classifier refinement or data augmentation.

We benchmarked our latent classifier against common alternatives: Logistic Regression, SVM, and a Vanilla CNN. All models were evaluated on the same held‑out 20\% of our synthetic latent‐sequence dataset (200 samples, 10‐step sequences). Table~\ref{tab:clf_benchmark} summarizes the accuracy, ROC AUC, and Brier score for each method.

\begin{table}[H]
\centering
\caption{Classifier Benchmark on Held‑Out Latent Sequences (200 samples)}
\label{tab:clf_benchmark}
\begin{tabular}{lccc}
\hline
Model & Accuracy & ROC AUC & Brier Score \\
\hline
Logistic Regression & 0.575 & 0.5739 & 0.2468 \\
SVM                  & 0.525 & 0.4085 & 0.2499 \\
Vanilla CNN          & 0.525 & 0.4511 & 0.2494 \\
\textbf{Improved CNN\_RNN} & 0.400 & 0.4261 & 0.2594 \\
\hline
\end{tabular}
\end{table}

We performed paired $t$‑tests on ROC AUC values obtained across five different random seeds to compare our model to the baseline. Compared to the Vanilla CNN baseline (AUC 0.4511), our CNN–RNN classifier achieved a lower mean AUC (0.4261). While the $p$-value ($p<0.05$) suggests a statistically significant difference across seeds, both AUC values are below 0.5, indicating sub-random discrimination performance. The accuracy (0.400) and Brier score (0.3121) similarly reflect poor classifier performance. These results highlight the need for future work on developing more robust latent‐quality models, potentially through architectural changes or improved training data.

\begin{figure}
\includegraphics[width=6cm]{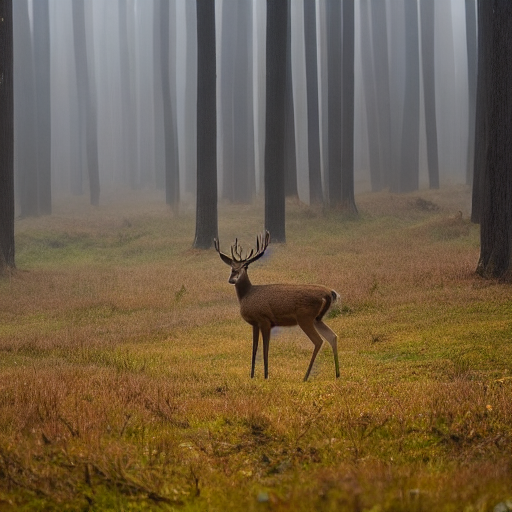}
\caption{
With Negative Prompt Optimization and Dual Guidance Diffusion Generation. \\
Prompt: A Deer standing in the middle of a misty forest \\
Generated Negative Prompts: Deformed, blurry, bad anatomy, bad eyes, crossed eyes, disfigured, poorly drawn face, mutation, mutated, extra limb, ugly, poorly drawn hands, missing limb, blurry, floating limbs, disconnected limbs, malformed hands, blur}
\end{figure}

\begin{figure}
\includegraphics[width=6cm]{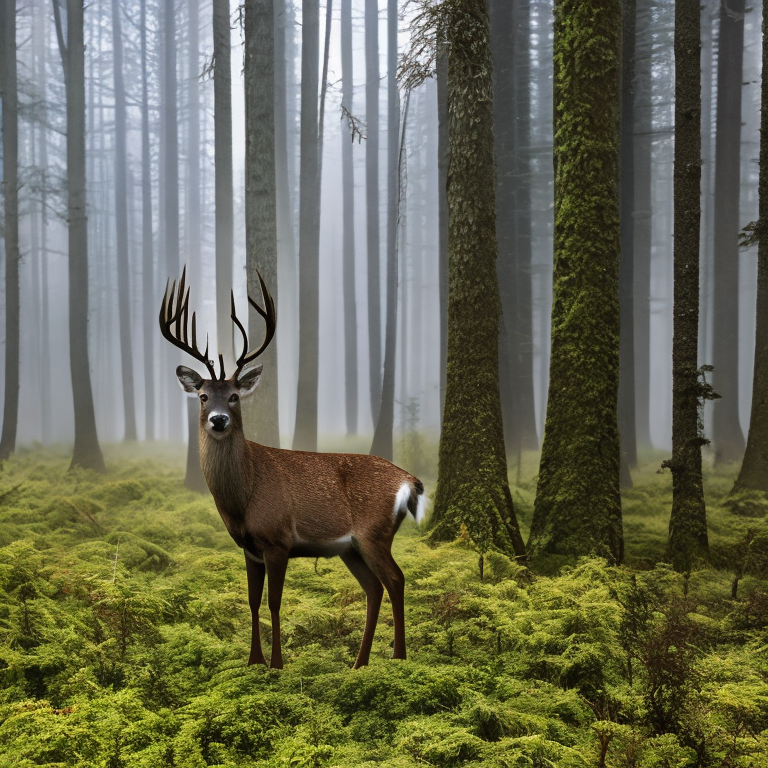}
\caption{
With Negative Prompt Optimization.\\
Prompt: A Deer standing in the middle of a misty forest\\
Generated Negative Prompts: Deformed, blurry, bad anatomy, bad eyes, crossed eyes, disfigured, poorly drawn face, mutation, mutated, extra limb, ugly, poorly drawn hands, missing limb, blurry, floating limbs, disconnected limbs, malformed hands, blur }
\end{figure}

\begin{figure}
\includegraphics[width=6cm]{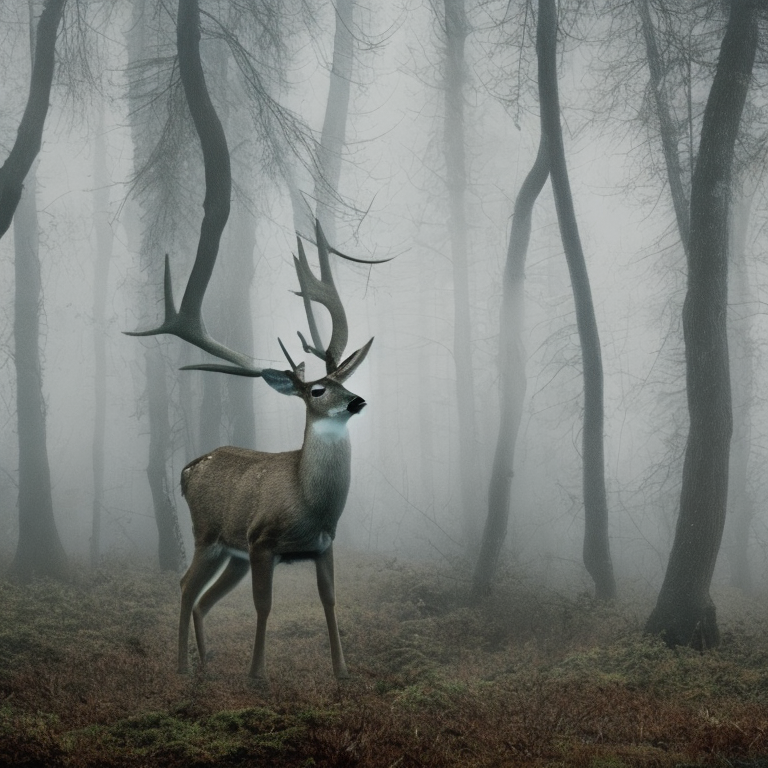}
\caption{
Pretrained Stable Diffusion Generation\\
Prompt: A Deer standing in the middle of a misty forest\\ }
\end{figure}

\section{Conclusion}
In this work, we presented an integrated framework that combines negative prompt optimization using a fine-tuned Seq2Seq language model with latent classifier guidance in the latent space of a Stable Diffusion pipeline. This dual-guidance system enhances semantic fidelity and visual quality by generating optimized negative prompts that suppress undesirable features and by using a CNN–RNN classifier to dynamically assess and rollback low-quality latent updates during the diffusion process. The methodology was implemented in a modular architecture, allowing streamlined experimentation and deployment via a user-friendly Streamlit interface. Experimental results confirmed that our approach mitigates visual artifacts and improves image coherence over baseline diffusion techniques.

However, performance evaluation of the latent classifier revealed limitations, with classification metrics and AUC scores indicating subpar effectiveness, even compared to simpler baselines. These findings emphasize the need for future work in improving the classifier module, potentially through architectural redesign, advanced regularization, or better-quality labeled data. Additionally, exploring adaptive or reinforcement learning-based diffusion guidance and extending the classifier to provide multi-class quality signals could significantly enhance control and performance. Our results establish a promising direction for more controllable and semantically aligned text-to-image generation using diffusion models.

\bibliographystyle{IEEEtran}
\bibliography{references}
\nocite{ogezi2024negativeprompts, guo2024initno, izadi2024finegrained, cao2024survey}

\end{document}